\documentclass[conference,letterpaper]{IEEEtran}
\IEEEoverridecommandlockouts

\usepackage{geometry}

\usepackage{cite}
\usepackage{svg}
\usepackage{amsmath,amssymb,amsfonts}
\usepackage{algorithm}
\usepackage{algorithmic}
\usepackage{graphicx}
\usepackage{textcomp}
\usepackage{xcolor}
\usepackage[numbers]{natbib}
\usepackage{booktabs}
\usepackage{multirow}
\usepackage{pdfpages}
\begin{document}

\title{Federated Learning Model Aggregation in Heterogenous Aerial and Space Networks}

\author{
\IEEEauthorblockN{Fan Dong\IEEEauthorrefmark{1}, Ali Abbasi\IEEEauthorrefmark{1}, Henry Leung\IEEEauthorrefmark{1}, Xin Wang\IEEEauthorrefmark{2}, Jiayu Zhou\IEEEauthorrefmark{3}, Steve Drew\IEEEauthorrefmark{1}}
\IEEEauthorblockA{
\IEEEauthorrefmark{1}Department of Electrical and Software Engineering, University of Calgary, Calgary, AB, Canada \\
\IEEEauthorrefmark{2}Department of Geomatics Engineering, University of Calgary, Calgary, AB, Canada \\
\IEEEauthorrefmark{3}Department of Computer Science and Engineering, Michigan State University, East Lansing, MI, USA \\
\{fan.dong, ali.abbasi1, leungh, xcwang\}@ucalgary.ca, zhou@cse.msu.edu, steve.drew@ucalgary.ca}
}

\maketitle

\begin{abstract}
Federated learning offers a promising approach under the constraints of networking and data privacy constraints in aerial and space networks (ASNs), utilizing large-scale private edge data from drones, balloons, and satellites.
Existing research has extensively studied the optimization of the learning process, computing efficiency, and communication overhead. 
An important yet often overlooked aspect is that participants contribute predictive knowledge with varying diversity of knowledge, affecting the quality of the learned federated models.
%
%A representative and widely adopted federated learning approach is FedAvg, which randomly selects clients and assigns weight based on the number of samples.
%However, such a weighting strategy does not consider the diversity of samples on each client, which could greatly affect local update performance and the final aggregated model.
%
In this paper, we propose a novel approach to address this issue by introducing a \texttt{Wei}ghted \texttt{Av}era\texttt{g}ing and \texttt{C}lient \texttt{S}election (\texttt{WeiAvgCS}) framework that emphasizes updates from high-diversity clients and diminishes the influence of those from low-diversity clients. 
Direct sharing of the data distribution may be prohibitive due to the additional private information that is sent from the clients.  
As such, we introduce an estimation for the diversity using a \textit{projection}-based method. 
Extensive experiments have been performed to show \texttt{WeiAvgCS}'s effectiveness. \texttt{WeiAvgCS} could converge 46\% faster on FashionMNIST and 38\% faster on CIFAR10 than its benchmarks on average in our experiments.

\end{abstract}


\section{Introduction}

%FL is emerging.
% Challenges in data heterogeneity and diversity.
% Existing methods
% How we approach this problem.
% Our contributions

Aerial and space networks (ASNs) \cite{liu2018space} are new types of networks providing networked integration of aerial and space assets. Drones, balloons, and satellites are connected together, collecting and relaying diverse sensing data with varying-speed and universal data networks. Additionally, the size of these assets in ASNs can provide edge computing capabilities \cite{zhang2022aerial} to perform complex machine-learning tasks for localized data processing. 
The nature of diverse types of devices, limited bandwidths, and different ownerships have posed challenges in processing data and using machine learning to train predictive models centrally in ASNs. These challenges arise because 1) the limited bandwidths may result in the transmission of massive amounts of data infeasible; 2) Due to privacy concerns and regulations, data cannot be permitted to be transmitted among ASN nodes; 3) The scheme of training ML models offloaded in a single place may introduce additional latency and power consumption. These resources are considered scarce in most of the nodes in ASNs.
% In the past decade, Artificial Intelligence (AI) has seen a tremendous impact on our daily lives thanks to the massive volume of data generated by the Internet of Things (IoT) and edge \cite{bonawitz2019towards,abreha2022federated} devices. 
% A typical centralized machine learning system collects and trains data on a centralized server. 
% The sharing of raw data may have significant privacy concerns in many applications.
% There has been growing recognition and increasing legislation efforts in various jurisdictions to protect the privacy of their residents, e.g., GDPR \cite{GDPR}. % and PIPEDA \cite{pipeda2021} in Canada.

\begin{figure}[htbp]
    \centering
    \includegraphics[width=0.95\linewidth]{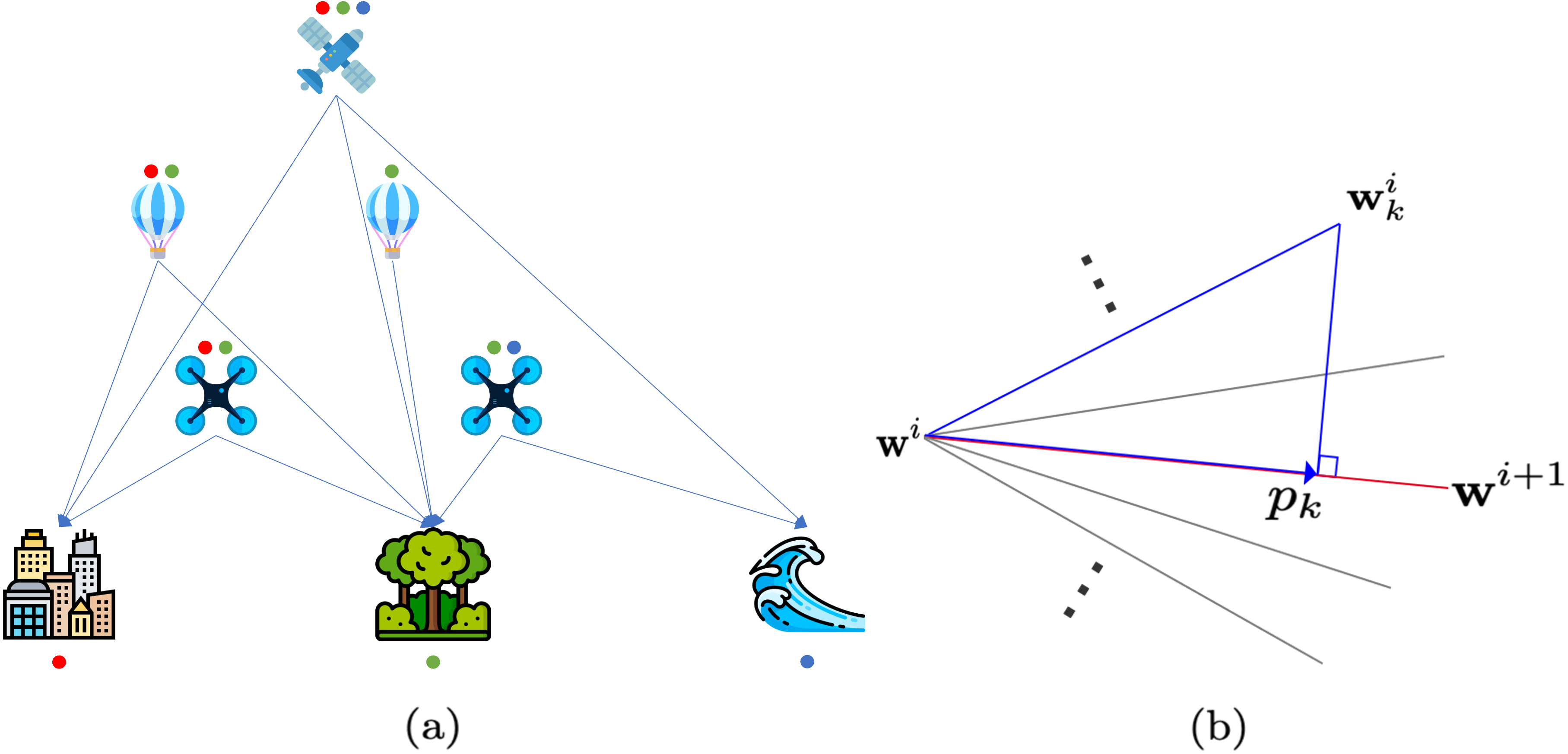}
    \caption{
    (a) Different devices would possess varying diversity levels of data concerning their locations, including orbiting heights and cruising directions. When performing the FL aggregation, we propose to assign higher weights to clients with more diverse data.
    (b) The \textit{projection} of client $k$'s local update  ($\textbf{w}^i$ to $\textbf{w}_k^i$) at the $i$th round onto the global update ($\textbf{w}^i$ to $\textbf{w}^{i+1}$).}
    \label{fig:diversity_and_projection}
\end{figure}

% \begin{figure}[t]
%     \centering
%     \includegraphics[width=0.9\linewidth]{pics/diversity_projection.png}
%     \caption{
%     (a) Two clients, A and B, with the same number of samples but different diversity. Intuitively, we expect client B to contribute more to the global model. 
%     (b) The \textit{projection} of client $k$'s local update  ($\textbf{w}^i$ to $\textbf{w}_k^i$) at the $i$th round onto the global update ($\textbf{w}^i$ to $\textbf{w}^{i+1}$).}
%     \label{fig:diversity_and_projection}
% \end{figure}

Federated learning (FL) \cite{mcmahan2017communication} has emerged as a promising solution where dsitributed clients train their models locally and send only the model parameters to the central server.
Despite its potential, FL still faces the challenges introduced by statistical and system heterogeneity \cite{bonawitz2019towards, wu2023fedle}, especially when applied to ASNs where heterogeneity patterns may change rapidly.
% Theoretical analysis \cite{li2019convergence} has proved that such heterogeneity would increase the convergence time and jeopardize the model performance.
ASN-based edge devices are designed to complete diverse tasks with various intensities, resulting in high data heterogeneity. For instance, satellites capture images of different parts of the earth in the satellite constellations for earth observation scenarios. When tackling a specific task, such heterogeneity would be reflected by diverse labels of oceans, lands, deserts, plains, cities, and countryside.
Several approaches have been proposed to mitigate the effects of statistical heterogeneity by introducing a proximal term to the objective function \cite{li2020federated}, adopting reinforcement learning \cite{wang2020optimizing} to counterbalance non-i.i.d data distribution, and accelerating the convergence of FL training by weighted averaging based on cosine similarity \cite{wu2021fast} of model gradients.
While existing studies tackled the challenges of data heterogeneity from various perspectives, the definition of ``data heterogeneity'' often omitted the heterogeneity in class diversity.

In this paper, we propose a \textit{projection}-based weighted averaging and client selection algorithm in FL to mitigate the impact of data heterogeneity more generically by utilizing the diversity of data. 
As an intuitive example shown in Fig. \ref{fig:diversity_and_projection} (a), different devices possess different data concerning their types, locations, and orbital heights. 
The arrows mean that devices have collected data from the targeting sources, like city (red dot), countryside (blue dot), and sea (blue dot). The colored dots above each device show the source diversity level of the data. 
Apart from the difference in the source, the diversity level from different sources would also vary according to specific tasks. Data from the urban area would be more diverse on human activities than those from the countryside or sea area. For animal activities, it should be the reversed situation. Naturally, when performing the aggregation, we tend to assign higher weights to clients with more diverse data and retain them to participate in more rounds of training.

% Given a dataset split and assigned to multiple clients, represented by the colored dots shown in Fig. \ref{fig:diversity_and_projection} (a), where each dot represents a data sample. 
% The color of a dot represents its respective label. 
% Suppose there are two clients, A and B, each assigned 9 samples. 
% Despite both clients having the same number of samples, Client B has significantly higher diversity. 
% Naturally, we tend to assign more weight to Client B when aggregating their model updates at the central server and retain Client B to participate in more rounds of training.
%
We quantify the diversity with the variance of label distribution. However, the requirement of uploading variance to the central server for weighted averaging and client selection may cost additional privacy of clients.  
% To quantify the degree of data diversity, the entropy of label distribution can be used, as entropy is defined by the average amount of information conveyed by an event \cite{jost2006entropy}. However, disclosing client data entropy to the server would compromise privacy. 
%
As an alternative, we propose a substitution metric that has a strong correlation with the label diversity, which is the \textit{projection}.
We then use the \textit{projection} as an approximation to perform the weighted averaging and client selection, as shown in Fig. \ref{fig:diversity_and_projection} (b). By this approach, we significantly improve the convergence speed of FL without asking for additional information. Furthermore, we prove that WeiAvgCS could be combined with other approaches like FedProx \cite{li2020federated}, MOON \cite{li2021model}, and Scaffold \cite{karimireddy2020scaffold} to further improve the performance. Our contributions are listed below:

\begin{itemize}
    \item We link the clients' data label diversity to the global model performance in FL. 
    % We verify this hypothesis by using the label distribution variance to represent the client data diversity. 
    To our knowledge, this is the first work to consider label diversity with weighted averaging and client selection in FL.
    \item We devise "\textit{projection}" as a highly correlated approximation of the label distribution variance to eliminate privacy concerns. 
    % The \textit{projection}-based metric proves to be highly correlated to label diversity represented by variance. 
    An algorithm called WeiAvgCS is proposed.
    % for \textit{projection}-based weighted averaging and client selection.
    \item We conduct extensive experiments to evaluate the performance of WeiAvgCS. The numerical results demonstrate the superiority of WeiAvgCS in various scenarios.
\end{itemize}

\section{Related Work}
In this section, we introduce three categories of existing studies relevant to our proposed method: previous schemes to mitigate the heterogeneity effect, different variants of weighted averaging, and other client selection approaches.

\subsection{Schemes to Mitigate the Heterogeneity Effect}

First proposed by Google \cite{mcmahan2017communication}, FL offered a promising paradigm to meet the conflicting demands of privacy protection and a vast volume of data for model training. The convergence time, communication cost, and privacy enhancement \cite{bonawitz2019towards, wu2023topology} remain critical research topics in FL. One key challenge in FL is the data heterogeneity among different clients. 
% In \cite{li2019convergence}, the negative impact of heterogeneity on FL has been theoretically analyzed. 

FedProx \cite{li2020federated} aimed to mitigate the statistical heterogeneity by introducing a proximal term in its optimization objective to regularize the training process, preventing the local training from over-fitting. 
MOON \cite{li2021model} utilized the similarity between model representations to correct the local training of individual parties to improve the model performance under heterogeneous distribution.
Scaffold \cite{karimireddy2020scaffold} used control variates to correct for the ‘client-drift’ in its local updates.
While adding additional regularization terms to local objective functions requires longer computation time, maintaining control variates also incurs additional communication costs.
% In \cite{INFOCOM2023_1}, an efficient knowledge transmission method was proposed for heterogeneous FL using a public dataset and weighted ensemble distillation scheme.

\subsection{Variants of Weighted Averaging}
There have also been studies for weighted model aggregation in FL. FedAT \cite{chai2020fedat} used a straggler-aware, weighted aggregation heuristic to steer and balance the training for further accuracy improvement. FedAdp \cite{wu2021fast} adopted the weighted averaging method to accelerate the convergence based on the angle between the local gradient and global gradient, which is only suitable for cross-silo situations.
In \cite{aamer2021entropy}, the authors adopted entropy based on clustering result as the weight for model aggregation, while they require all clients' participation and the entropy information to be uploaded to the central server, which may incur privacy violation problems.

\subsection{Client Selection Approaches}
A client selection method was adopted to counterbalance the bias introduced by non-i.i.d. data distribution \cite{wang2020optimizing}. POWER-OF-CHOICE selects clients with higher local losses to improve the convergence speed \cite{cho2020client}. FedCS \cite{nishio2019client} focuses on client selection under heterogeneous resources in mobile edge. However, they didn't consider the diversity differences among clients.

% To achieve better convergence performance in cross-device scenarios, we proposed WeiAvgCS based on the variance of label distribution. Furthermore, we demonstrate the strong correlation between \textit{projection} of local updates on global update and variance, making it possible to perform weighted averaging solely based on the local updates. We compare the performance of our design with some state-of-the-art approaches and their combined approaches, proving our algorithm's effectiveness.

\section{Problem Formulation}
The intuition of our proposed method can be sourced to acknowledge the benefits of diversity, systematically summarized in \cite{gong2019diversity}. To our knowledge, label diversity has not been widely studied in FL. 
% In FedAvg \cite{mcmahan2017communication}, the local updates are treated equally during model aggregation if they possess the same amount of samples. However, the quality of samples located on each client can be different. 
% We can verify our hypothesis if diversity can be quantified for each client.

% \subsection{Quantifying Diversity by Entropy}
\subsection{Quantifying Diversity by Variance}

To measure the class diversity of each client, we turn to the variance of the label distribution. For a dataset $X$, suppose there are $B$ kinds of classes. Denote the index of a client in FL by $i \in [1, C]$, where $C$ is the total number of participating clients. The label distribution on client $i$ can be represented by $\mathbf{p_i} = [p_{i, 1}, \cdots, p_{i, j}, \cdots, p_{i, B}]$, where $p_{i, j}$ denotes the portion of samples with $j$th class on client $i$. Intuitively, a higher variance of $\mathbf{p_i}$ indicates a skewed distribution and less diverse samples on a client. Therefore, we use the negative variance of $\mathbf{p_i}$ to evaluate the diversity of the dataset on client $i$.
\begin{equation}
d_i = - \sigma^2(\mathbf{p_i})    
\end{equation}

Based on this variance-based definition of label diversity, we will increase the impact of the high-diversity clients by assigning higher weights to them and retaining corresponding clients to participate in more rounds of training.

\subsection{Theoretical Analysis}
For the training dataset $X$, suppose there are $B$ kinds of classes. Then, $X$ can be split by classes: $X = \{X_{C_1}, X_{C_2}, \cdots, X_{C_{B}}\}$.
Generally, we want to train a model that generalizes best and treats each kind of class equally important. So our objective function for model training should also be evaluated in the same way. We assume each client contains the same amount of samples for simplification.

Based on the aforementioned idea, we can decompose the objective function, a.k.a. loss function, as the average loss on each class with an identical weight $\frac{1}{B}$.

\begin{equation}
    \label{GlobalLoss}
    F(X, \textbf{w}) = \sum_{j=1}^{B}\frac{1}{B} F(X_{C_j}, \textbf{w}).
\end{equation}
And the global gradient will be

\begin{equation}
    \label{GlobalGradient}
    \nabla F(X, \textbf{w}) = \sum_{j=1}^{B}\frac{1}{B}\nabla F(X_{C_j}, \textbf{w})
\end{equation}

In Equation (\ref{GlobalLoss}) and Equation (\ref{GlobalGradient}), $\textbf{w}$ denotes the model parameters. For simplicity, $\textbf{w}$ may be omitted.

In FL, we use the aggregation of the local gradients as an approximation of the centralized global optimum. However, there may be approximation errors because of the discrepancy between local and global objective functions. And the discrepancy is incurred by the distribution heterogeneity on local datasets. We use $\mathbf{p_i} = [p_{i, 1}, p_{i, 2}, \cdots, p_{i, B}]$ to measure the distribution of different classes on client $i$, where $\sum_{j=1}^{B}p_{i,j}=1$, Then the local gradient can be represented as

\begin{equation}
    \label{LocalGradient}
    \nabla F_i(X_i) = \sum_{j=1}^{B} p_{i,j} \nabla F(X_{i, C_{j}})
\end{equation}

If we parameterize the aggregation with different weight $\omega_i$ for different client $i$. Note that $\sum_{i=1}^{M}\omega_i = 1$. Then

\begin{equation}
    \label{GlobalGradientAproximation2}
    \begin{split}
        \widetilde{\nabla {F(x)}} 
        & =  \sum_{i=1}^{M} \omega_i \nabla F_i(X_i)\\
        & = \sum_{i=1}^{M} \omega_i \sum_{j=1}^{B} p_{i,j} \nabla F(X_{i, C_{j}}) \\
        & = \sum_{j=1}^{B} \sum_{i=1}^{M} \omega_i p_{i,j} \nabla F(X_{i, C_{j}})
    \end{split}
\end{equation}

The approximation error will be

\begin{equation}
    \label{GradientError2}
    \begin{split}
        &\parallel \nabla F(x) - \widetilde{\nabla {F(x)}} \parallel \\
        &= \Big\lVert \sum_{j=1}^{B} \Big ( \frac{1}{B}\nabla F(X_{i, C_j}) - \sum_{i=1}^{M} \omega_i p_{i,j} \nabla F(X_{i, C_{j}}) \Big ) \Big\rVert \\
        & \leq \sum_{j=1}^{B} \parallel \nabla F(X_{i, C_j})\parallel \Big|\frac{1}{B} - \sum_{i=1}^{M} \omega_i p_{i,j}\Big| 
    \end{split}
\end{equation}

For $p_{i,j}$, we can get that its expectation is $\frac{1}{B}$, and its variance is denoted by $\sigma_i^2$. Across all of the $M$ clients, we know the expectation and variance of $\sum_{i=1}^{M} \omega_i p_{i,j}$ are $\frac{1}{B}$ and  $\sum_{i=1}^{M}\omega_i^2 \sigma_i^2$.
Thus term $\frac{1}{B} - \sum_{i=1}^{M} \omega_i p_{i,j}$ has expectation of $0$ and variance of $\sum_{i=1}^{M}\omega_i^2 \sigma_i^2$.
Since a smaller variance means a tighter bound approximation error, we have two approaches to tighten the approximation error. (1) We distribute $\omega_i$ inversely related to $\sigma_i$ to reduce the overall variance. However, we don't want to assign all of the weight to the client with the least $\sigma_i$, which will lead other clients' updates to be completely unused. (2) In the client selection step, we select clients with higher class diversity. We can achieve this by retaining the high diversity of clients we selected in the previous round.

% Let $X$ be a discrete random variable such that $x \in \mathcal{X}$ and the probability density function 
% \begin{equation}
%     p(x) = Pr\{X = x\}, x \in \mathcal{X}
% \end{equation}
% The entropy of variable X is then defined by $H(X)$, where
% \begin{equation}
%     H(X) = -\sum_{x \in \mathcal{X}} p(x) \log p(x)
% \end{equation}
\section{The WeiAvgCS Algorithm}
To verify our hypothesis that clients with high-diversity classes improve the model performance, we propose an algorithm named \textit{WeiAvgCS}, shown in Algorithm \ref{alg:WeiAvgCS}.

\begin{algorithm}[H]
\caption{WeiAvgCS}\label{alg:WeiAvgCS}
\begin{algorithmic}[1]
        \REQUIRE Initial global model $\textbf{w}^0$, number of participating clients in each round $n$, number of global epochs $I$, maximum retaining rounds for each client $R$, number of clients retained each round $r$, total client pool $C$.
        \FOR{global epoch $i$ in $[1, 2, 3, \cdots, I]$}
            \IF{i==1}
            \STATE $\boldsymbol{c}$ = Randomly select $n$ clients $\boldsymbol{c}$ from $C$
            \ELSE
            \STATE Retain $r$ clients $\boldsymbol{c_r}$ with highest value based on $d_{i-1}$.
            \STATE Randomly select $n-r$ clients $\boldsymbol{c_{n-r}}$ from $C$.
            \STATE If any clients in $\boldsymbol{c_r}$ and $\boldsymbol{c_{n-r}}$ are selected R times right before the $i_{th}$ round, replace them with other clients.
            \STATE $\boldsymbol{c}$ = $\boldsymbol{c_r}$ + $\boldsymbol{c_{n-r}}$
            \ENDIF
            
            \STATE Send the current global model $\textbf{w}^{i}$ to $\boldsymbol{c}$. Perform local training and upload the local updates $\boldsymbol{u^{i}_\cdot} = [u^{i}_{1}, u^{i}_{2}, \cdots, u^{i}_{n}]$ to the central server.
            
            \IF{USE\_VARIANCE}
            \STATE Compute the diversity of each client in $\boldsymbol{c}$ as $\boldsymbol{d_i}$.
            \ELSE
            \STATE Simple average $\boldsymbol{u^{i}_\cdot}$ to obtain the temporal global update $\boldsymbol{u^{i+1}}$.
            \STATE For each local update of $\boldsymbol{u^{i}_\cdot}$, calculate its \textit{projection} on the temporal global update, which would yield a \textit{projection} vector $\boldsymbol{p_i}=\{p_{i, 1}, p_{i, 2}, \cdots, p_{i, n}\}$.
            \STATE Use $\boldsymbol{p_i}$ as the estimation of $\boldsymbol{d_i}$: $\boldsymbol{d_i}$ = $\boldsymbol{p_i}$.
            \ENDIF

            \STATE $\boldsymbol{z}$ $\gets$ $(d-min(d))/(max(d)-min(d))$.
            \STATE $\boldsymbol{z'}$ $\gets$ $\boldsymbol{(z + 1)}^{\lambda}$.
            \STATE Calculate the weight of each client in $\boldsymbol{c}$: $\boldsymbol{w}=\boldsymbol{z'}/\sum \boldsymbol{z'}$.
            \STATE Conduct the weighted averaging to obtain the weighted global update: $\boldsymbol{U^{i}}=\sum(\boldsymbol{w} \cdot \boldsymbol{u^i_{\cdot}})$.
            \STATE Update the global model: $\textbf{w}^{i+1} \gets \textbf{w}^{i} + \boldsymbol{U^{i}}$.
        \ENDFOR
        \RETURN $\textbf{w}^I$
\end{algorithmic}
\end{algorithm}

This algorithm sets the initial global model as $\textbf{w}^0$. Suppose there are $N$ clients in total. In each round, $n$ clients are selected for training. The training will last a total of $I$ rounds. 

From the second round, we start to retain high-diversity clients from the previous round to participate in the next epoch of training. Hyperparameters $r$ and $R$ will control how many clients will be retained and how many rounds each client participates in the training in a row respectively. The client selection part is shown in Algorithm \ref{alg:WeiAvgCS} from line 2 to line 9.

For some scenarios, we can ask clients to upload their label diversity to the central server. However, this would not be always practical. So we need to estimate the client diversity then. The client diversity calculation and estimation are shown in Algorithm \ref{alg:WeiAvgCS} from line 11 to line 17. In the next subsection, we will illustrate the details of the estimation of diversity when uploading diversity to the central server is not available.
In line 18 of Algorithm \ref{alg:WeiAvgCS}, we perform a min-max normalization to the (estimated) diversity vector.
In line 19 Algorithm \ref{alg:WeiAvgCS}, we apply a non-linear transformation to control the weights assigned to clients with different diversity. 
Hyperparameter $\lambda$ controls the degree of emphasis given to updates from high-diversity clients.
When $\lambda$ equals 0, WeiAvgCS will degenerate to FedAvg.
We distribute the same amount of samples to each client for simplification.

\subsection{Using \textit{Projection} As An Approximation}

Despite that WeiAvgCS based on variance could improve the performance compared to FedAvg. Uploading clients' label distribution variance might cause privacy concerns.
% , as label distribution variance would expose information on data distribution.

To address this challenge, we propose a metric called \textit{projection} to estimate the label distribution variance. The calculation of \textit{projection} is shown in Fig. \ref{fig:diversity_and_projection} (b), where $\textbf{w}^i$ represents the flattened global model parameters of the $i_{th}$ round before sending to clients, $\textbf{w}_k^i$ represents the updated flatten model parameters on client $k$, and $\textbf{w}^{i+1}$ represents the simple averaged global model after clients send their local model updates to the central server, note that $\textbf{w}^{i+1}$ is the same with the aggregated model in FedAvg. The \textit{projection} norm shown in Fig. \ref{fig:diversity_and_projection} (b) could be calculated as:
\begin{equation}
p_k = \frac{(\textbf{w}_k^i - \textbf{w}^i) \cdot (\textbf{w}^{i+1} - \textbf{w}^i) }{\lVert \textbf{w}^{i+1} - \textbf{w}^i \rVert}
\end{equation}

\begin{figure}[htbp]
    \centering
    \includegraphics[width=0.75\linewidth]{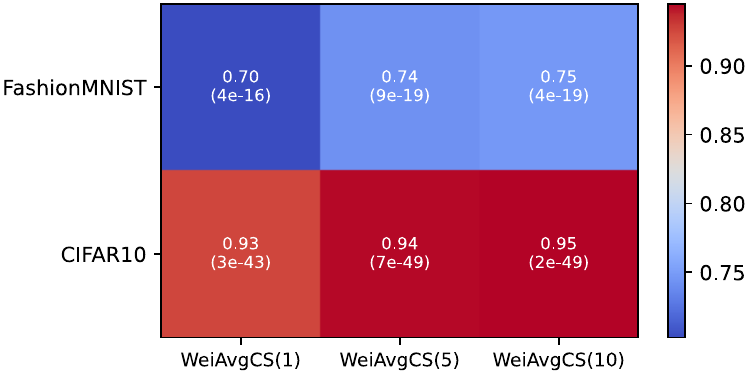}
    \caption{Correlation between \textit{projection} and variance of WeiAvgCS with different $\lambda$ on FashionMNIST and CIFAR10.}
    \label{fig:heatmap}
\end{figure}

We build a correlation heatmap as illustrated in Fig. \ref{fig:heatmap}, wherein for each grid, we concatenate all \textit{projection} values in an epoch and average them based on the client ID. Then, we calculate the Pearson correlation coefficient and the p-value. 
We performed such calculations on each dataset as shown in each row in Fig. \ref{fig:heatmap}. Within each dataset, we also applied different $\lambda$ in WeiAvgCS to further test the correlation between \textit{projection} and variance, as shown in each column in Fig. \ref{fig:heatmap}. The number in the parentheses of horizontal axis denotes the value of $\lambda$. The number in each grid represents the correlation coefficient (above) and the corresponding p-value (below). The results show a strong positive correlation, showing the promise of using the \textit{projection} values as an estimation of variance to perform weighted averaging and client selection.

\section{Experiment Results}

We conducted our experiments on FashionMNIST and CIFAR10.
For both FashionMNIST and CIFAR10, we distribute $500$ samples across a total of $100$ clients. 
To better control the diversity distribution, we devised a manual distribution scheme with a hyperparameter $p$ to control the label variance distribution, instead of the Dirichlet distribution widely used in other studies \cite{cho2020client, yu2023latency}.
% The variance of label distribution among $100$ clients is controlled by a hyperparameter $p$. 
% The distribution could be seen as Fig. \ref{user_entropy} when $p=1$, where $N=100$. 
We will discuss the effect of $p$ on our algorithm later.
% \begin{figure}[htbp]
%     \centering
%     \includesvg[width=0.8\linewidth]{pics/user_label_diversity}
%     \caption{Distribution of variance of each user's label distribution.}
%     \label{user_entropy}
% \end{figure}
We select $n=10$ clients in each global epoch to perform local training. During local training, the batch size is set to be $B=32$, the learning rate is set to be $lr=0.01$, and the number of local epochs is $E=10$.
We apply the same ``sgd'' optimizer with a momentum of 0.9 for each dataset. The L2 normalization coefficient is set as 1e-4.
% The configurations for each dataset are summarized in Table \ref{hyperparameter_table}. 
Due to the stochastic nature of FL in the client selection step, we employ hundreds of different random seeds to obtain an average performance for comparison in the following experiment. This approach not only ensures a more consistent evaluation but also enables us to draw conclusions with statistical confidence.

% \begin{table}[htbp]
% \centering
% \caption{Hyperparameters for each dataset.}
% \begin{tabular}{c|ccccc}
% \hline
% & $M$ & $N$ & $n$ & $B$ & $E$ \\
% \hline
% FashionMNIST & 500 & 100 & 10 & 32 & 10 \\
% CIFAR10 & 500 & 100 & 10 & 32 & 10 \\
% \hline
% \end{tabular}
% \label{hyperparameter_table}
% \end{table}

\begin{figure}[htbp]
    \centering
    \includegraphics[width=0.9\linewidth]{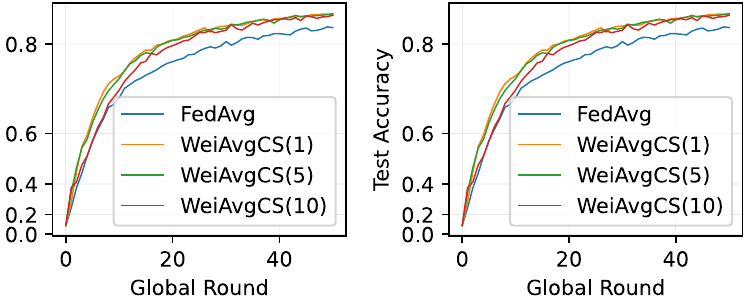}
    \caption{WeiAvgCS($\lambda$) with using (a) variance and (b) projection.}
    \label{Tune_WeiAvgCS_Lambda}
\end{figure}

\subsection{WeiAvgCS with different $\lambda$}
As shown in Fig. \ref{Tune_WeiAvgCS_Lambda} (a), we performed WeiAvgCS using variance with different $\lambda$ on FashionMNIST. WeiAvgCS with a wide range of $\lambda$ shows superiority over vanilla FedAvg. Large $\lambda$ like 10 may introduce a large advantage initially while causing instability in the later epochs. Later, we used \textit{projection} as an approximation of variance to perform the same experiment as shown in Fig. \ref{Tune_WeiAvgCS_Lambda} (b). The accuracy line shows a similar pattern with Fig. \ref{Tune_WeiAvgCS_Lambda} (a). This proved that our algorithm works and could achieve better convergence performance over FedAvg on a wide range of $\lambda$ choices.

\begin{figure*}[htbp]
    \centering
    \includegraphics[width=0.87\linewidth]{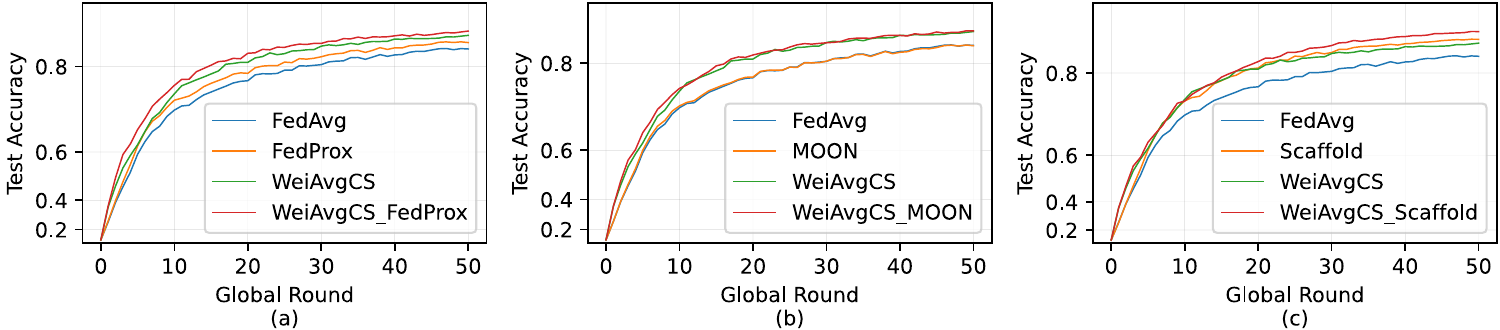}
    \caption{Test accuracy of FedAvg, WeiAvgCS with 3 benchmarks (a) FedProx, (b) MOON, (c) Scaffold and their corresponding combination on FashionMNIST(p=1).}
    \label{FashionMNIST}
\end{figure*}

\begin{figure*}[htbp]
    \centering
    \includegraphics[width=0.87\linewidth]{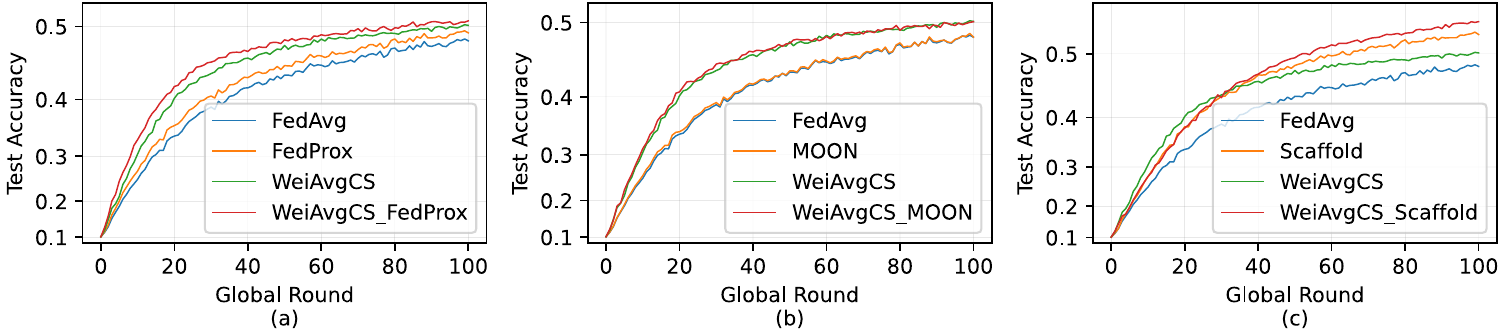}
    \caption{Test accuracy of FedAvg, WeiAvgCS with 3 benchmarks (a) FedProx, (b) MOON, (c) Scaffold and their corresponding combination on CIFAR10(p=1).}
    \label{cifar10}
\end{figure*}

\begin{table*}[htbp]
\centering
\caption{Experiment results for FashionMNIST and CIFAR10 under different distribution.}
\begin{tabular}{ccccccccc}
\toprule
                   & \multicolumn{4}{c}{FashionMNIST}                        & \multicolumn{4}{c}{CIFAR10}                       \\ \hline
                   & \multicolumn{2}{c}{p=1} & \multicolumn{2}{c}{p=0} & \multicolumn{2}{c}{p=1} & \multicolumn{2}{c}{p=0} \\ \hline
                   & Rounds & TestAcc(std)   & Rounds & TestAcc(std)   & Rounds & TestAcc(std)   & Rounds & TestAcc(std)   \\ \hline
FedAvg             & 46     & 0.8291(0.0332) & 50     & 0.8542(0.0207) & 97     & 0.4822(0.0400) & 90     & 0.5177(0.0294) \\
FedProx            & 38     & 0.8391(0.0282) & 39     & 0.8598(0.0170) & 79     & 0.4922(0.0354) & 82     & 0.5215(0.0270) \\
MOON               & 46     & 0.8286(0.0528) & 47     & 0.8551(0.0206) & 88     & 0.4822(0.0395) & 90     & 0.5161(0.0290) \\
Scaffold           & 25     & 0.8559(0.0637) & 25     & 0.8756(0.0085) & 50     & 0.5259(0.0294) & 44     & 0.5627(0.0245) \\
WeiAvgCS           & 30     & 0.8500(0.0201) & 31     & 0.8659(0.0127) & 58     & 0.5013(0.0298) & 67     & 0.5268(0.0179) \\
WeiAvgCS\_FedProx  & 24     & 0.8563(0.0158) & 28     & 0.8686(0.0129) & 50     & 0.5067(0.0238) & 62     & 0.5258(0.0195) \\
WeiAvgCS\_MOON     & 26     & 0.8508(0.0226) & 30     & 0.8686(0.0115) & 60     & 0.5010(0.0290) & 65     & 0.5257(0.0140) \\
WeiAvgCS\_Scaffold & 23     & 0.8674(0.0098) & 24     & 0.8784(0.0052) & 45     & 0.5421(0.0653) & 41     & 0.5795(0.0131) \\ \bottomrule
\end{tabular}
\label{experiment_table}
\end{table*}

\subsection{Comparing WeiAvgCS with Different Benchmarks.}

To prove WeiAvgCS's performance, we implemented some state-of-the-art benchmarks (FedProx \cite{li2020federated}, MOON \cite{li2021model}, Scaffold \cite{karimireddy2020scaffold}) proposed in FL to address the heterogeneity problem. Since WeiAvgCS is orthogonal to these algorithms, we also combine them with WeiAvgCS together to see if further improvement could be achieved. 
As shown in Fig. \ref{FashionMNIST} and Fig. \ref{cifar10}, The experiments on FashionMNIST and CIFAR10 shows WeiAvgCS's superiority. Compared with FedProx, WeiAvgCS could achieve better performance improvement both on FashionMNIST and CIFAR10. After the combination, the performance could be enhanced further. For  MOON, WeiAvgCS could also improve the performance of the combined approach even if MOON is barely working in our scenario. For Scaffold, WeiAvgCS could only achieve better convergence performance in the beginning rounds. After combining WeiAvgCS and Scaffold together, the combined approach could achieve significantly better performance than both Scaffold and WeiAvgCS.

We also conduct the same experiments with different distribution settings with p being 0. When p is small, the clients' diversity distribution would be more even regarding different degrees of diversity. The experiment results can be seen in Table. \ref{experiment_table}. Column ``rounds" means how many rounds each algorithm needs to reach the target accuracy. The target accuracy is determined by the lowest accuracy each algorithm could achieve. Column ``TestAcc(std)" means the final accuracy after a certain number of global rounds and the standard error that final accuracy among different random seeds.

\subsection{Identifying Limitations: Cases When WeiAvgCS Falls Short}
\begin{figure}[htbp]
    \centering
    \includegraphics[width=0.8\linewidth]{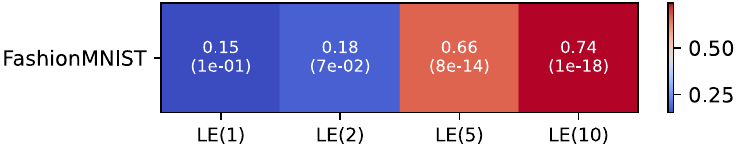}
    \caption{Correlation between \textit{projection} and variance with different local training epochs (1, 2, 5, and 10). Higher correlations are observed with a higher number of local epochs.}
    \label{heatmap_NotWorking}
\end{figure}

\begin{figure}[htbp]
    \centering
    \includegraphics[width=0.75\linewidth]{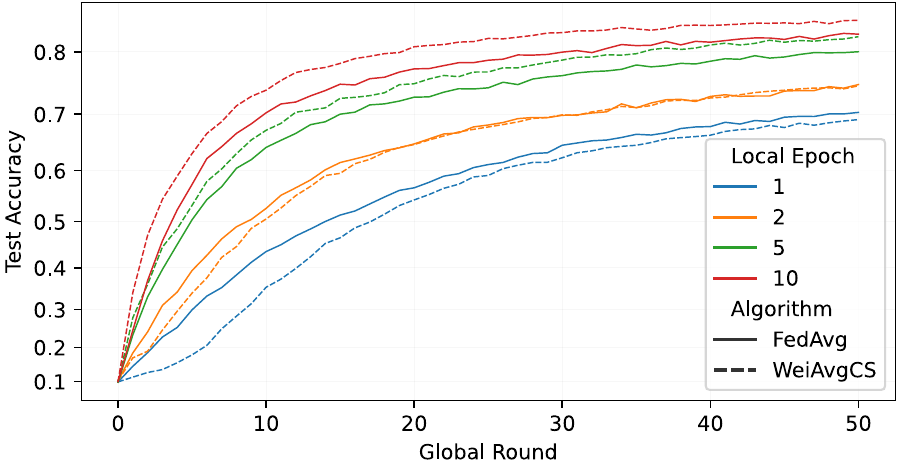}
    \caption{Convergence line with different local training epochs. WeiAvgCS shows no improved performance with too small local epochs.}
    \label{DifferentLocalEpochs}
\end{figure}

The WeiAvgCS algorithm works conditional on the strong correlation of \textit{projection} with variance. When we decrease the local epochs or by any means cause the local model under-fitted a lot, the correlation may decrease, which will lead to the WeiAvgCS failing. The experimental results on FashionMNIST, as depicted in Fig. \ref{heatmap_NotWorking}, demonstrate the effect of under-fitting. It can be observed that reducing local epochs to 1 or 2 weakens the correlation. Consequently, in Fig. \ref{DifferentLocalEpochs}, WeiAvgCS tends to exhibit poorer performance than FedAvg.
Conversely, when the number of local epochs is increased to 5 or 10, the correlation becomes stronger, leading to enhanced convergence performance for WeiAvgCS.

\section{Conclusion}
This paper introduces Weighted Federated Averaging and Client Selection (WeiAvgCS), a novel approach that addresses the issue of data heterogeneity and diversity in FL. By assigning weights based on the diversity of label distribution and retaining clients with high diversity in the previous round, WeiAvgCS emphasizes updates from high-diversity clients and diminishes the influence of low-diversity clients by utilizing a projection-based approximation method to estimate client data diversity without compromising individual client information. 
% This ensures that WeiAvgCS based on diversity can be performed while preserving privacy. 
%
The entire calculation would be performed on the central server and is based only on the model updates. Therefore, no additional communication costs would be incurred. The computational cost in the central server would also be negligible because of the low complexity.
Additionally, we conducted extensive experiments to show the effectiveness of WeiAvgCS. The orthogonality of WeiAvgCS with other state-of-the-art algorithms proved that WeiAvgCS could improve the convergence performance even further based on other algorithms.

\bibliography{ref}

\end{document}